\documentclass[twoside,11pt]{article}

\usepackage{blindtext}

\usepackage[abbrvbib, preprint]{jmlr2e}

\usepackage{mathtools}

\usepackage{bm}
\usepackage{amsbsy,mathptmx}

\usepackage{lastpage}
\usepackage[cal = cm, scr = dutchcal]{mathalpha}

\usepackage[table]{xcolor}
\definecolor{ao}{rgb}{0.0, 0.5, 0.0}

\usepackage{subcaption}

\usepackage{pifont, upgreek, textgreek, xparse}
\DeclareSymbolFont{Symbols}{OMS}{zplm}{m}{n}
\DeclareMathSymbol{\Infty}{\mathord}{Symbols}{"31}

\usepackage{multicol}
\usepackage{enumitem}
\usepackage{booktabs}
\usepackage{algorithm,algorithmic}
\usepackage{caption}
\def\argmax{\mathop{\rm argmax}}
\def\argmin{\mathop{\rm argmin}}

\newcommand \algocm[1]{\textcolor{violet}{\footnotesize\ttfamily /* \ #1 */}}

\NewDocumentCommand{\grad}{e{_^}}{%
  \mathop{}\!
  \nabla
  \IfValueT{#1}{_{\!#1}}
  \IfValueT{#2}{^{#2}}
}

\ShortHeadings{\textsc{BVSIMC}}{Fan et al.}
\firstpageno{1}

\begin{document}

\title{BVSIMC: Bayesian Variable Selection-Guided Inductive Matrix Completion for Improved and Interpretable Drug Discovery}

\author{\name Sijian Fan \email sfan@email.sc.edu \\
       \addr Department of Statistics\\
       University of South Carolina\\
       \name Liyan Xiong \email liyan@email.sc.edu \\
       \addr Department of Biostatistics\\
       University of South Carolina\\
       \name Dayuan Wang \email dayuan.wang@ufl.edu \\
       \addr Department of Biostatistics\\
       University of Florida\\
       \name Guoshuai Cai \email guoshuai.cai@surgery.ufl.edu \\
       \addr Department of Surgery \\
       University of Florida\\
       \name Ray Bai \email rbai2@gmu.edu \\
       \addr Department of Statistics\\
       George Mason University}


\editor{My editor}

\maketitle

\begin{abstract}
  Recent advances in drug discovery have demonstrated that incorporating side information (e.g., chemical properties about drugs and genomic information about diseases) often greatly improves prediction performance. However, these side features can vary widely in relevance and are often noisy and high-dimensional. We propose 
 \textit{Bayesian Variable Selection-Guided Inductive Matrix Completion} (BVSIMC), a new Bayesian model that enables variable selection from side features in drug discovery. By learning sparse latent embeddings, BVSIMC improves both predictive accuracy and interpretability. We validate our method through simulation studies and two drug discovery applications: 1) prediction of drug resistance in Mycobacterium tuberculosis, and 2) prediction of new drug-disease associations in computational drug repositioning. On both synthetic and real data, BVSIMC outperforms several other state-of-the-art methods in terms of prediction. In our two real examples, BVSIMC further reveals the most clinically meaningful side features.
\end{abstract}

\begin{keywords}
  spike-and-slab group lasso, inductive matrix completion, Bayesian variable selection, drug discovery
\end{keywords}

\section{Introduction}

\textit{De novo} drug development typically takes 10 to 17 years, with a less than 10\% probability of success \citep{ashburn2004drug}. Drug discovery, or the process of identifying new potential medications for disease targets, is crucial for mitigating costs and risk of failure for the pharmaceutical and biotechnology industries. While there are numerous aspects of drug discovery, two particularly pertinent issues are drug resistance and repositioning. 

Drug resistance can render developed drugs useless, directly contributing to the high cost of drug development \citep{Xia2017Bioinformatics}. Drug resistance is a leading cause of therapeutic failure in oncology and infectious disease \citep{Branda2024, ChenDrugResistance2019}. Notable examples include bacterial resistance to penicillin and antiretroviral resistance to HIV-1 \citep{Xia2017Bioinformatics}. Consequently, drug resistance prediction plays a critical role in drug discovery \citep{Xia2017Bioinformatics, Leighow2020MultiScale}. Predicting drug resistance during preclinical studies can guide the design of more effective drugs, e.g., optimization of drug potency \citep{Carbonell2014, Leighow2020MultiScale}. It can also improve clinical trial success rates by helping researchers determine whether to continue or discontinue the development of drug candidates \citep{Sommer2017nature}.

Because traditional \textit{de novo} drug discovery is often time-consuming and risky, drug repositioning has emerged as a cost-efficient and rapid approach to drug discovery. Drug repositioning (also known as redirecting, repurposing, or reprofiling) is defined as finding new indications and therapeutic uses for existing drugs. \citet{ashburn2004drug} and \citet{dudley2011exploiting} give three prominent examples of drug repositioning. Atomoxetine was originally developed for Parkinson’s disease but was later used to treat attention deficit hyperactivity disorder (ADHD). Minoxidil was originally developed for hypertension but was later used to treat male pattern hair loss. Finally, Viagra was originally used for angina but was later repositioned to treat erectile dysfunction and pulmonary hypertension. Compared with \textit{de novo} drug discovery, drug repositioning dramatically reduces the time and cost of drug development and alleviates concerns about safety and pharmacokinetic uncertainty \citep{ashburn2004drug}.

Both drug resistance prediction and drug repositioning can be represented by a binary matrix. Here, the rows of the matrix represent drugs, the columns represent diseases, and the individual elements represent drug-disease interactions (i.e., a disease resistance to a drug or a drug-disease association). In particular, if the $(i,j)$th element of the matrix is a ``1,'' this means that there is a known interaction between the $i$th drug and the $j$th disease. On the other hand, an entry of ``0'' could mean either that an interaction is known to be absent \emph{or} that it is unknown. In general, entries of ``1'' are considered to be more trustworthy since they have been manually verified \citep{liu2016neighborhood}. 

As many of the interactions in drug resistance prediction and drug repositioning are unknown, it is of interest to predict \emph{new} drug-disease interactions. In these two problems, there is often side information, or meta-data, which can enhance the prediction of drug-disease interactions \citep{zhang2020drimc,burkina2021inductive}. Side information is defined as additional data on the drugs (e.g., chemical structures) or the diseases (e.g., gene mutations). When side features are used to predict the entries in a data matrix, we call this \textit{inductive matrix completion} (IMC). In recent years, considerable effort has been devoted to leveraging side information in drug discovery. \citet{liu2016neighborhood} proposed neighborhood regularized logistic matrix factorization (NRLMF), which uses Laplacian regularization to ensure that drugs with similar chemical structures and diseases with similar genomic profiles have similar latent representations. \citet{zhang2020drimc} introduced a Bayesian IMC model for drug repositioning called DRIMC. Specifically, \citet{zhang2020drimc} fused different sources of side information into a similarity matrix for drugs and a similarity matrix for diseases and then used these similarity matrices as new side information matrices in an IMC model. However, both NRLMF and DRIMC utilize all of available the side information and may not perform well when side features are noisy and/or irrelevant.

Side information is often high-dimensional and noisy. For example, genomic side features are typically very high-dimensional, with few genes that are relevant for predicting disease resistance to a particular drug \citep{burkina2021inductive}. Noisy, high-dimensional side information can degrade predictive performance \citep{chiang2015matrix, burkina2021inductive}. To address this problem, \citet{chiang2015matrix} proposed DirtyIMC, which balances the observed data and side features and applies trace norm regularization to prevent overfitting. \citet{burkina2021inductive} later proposed sparse group IMC (SGIMC), which regularizes the side features through $\ell_{2,1}$ penalties. Both DirtyIMC and SGIMC are frequentist (non-Bayesian) approaches which apply the same amount of regularization to all side features. This can make it difficult to determine which side features are actually relevant for predicting drug-disease interactions. 

Motivated by the limitations of existing methods, we introduce BVSIMC, a new Bayesian IMC model guided by Bayesian variable selection. Specifically, we use spike-and-slab priors to shrink negligible side feature effects to zero. Unlike NRLMF and DRIMC, BVSIMC performs variable selection from the side features, thus filtering out side information which is redundant or unhelpful. Unlike DirtyIMC and SGIMC, BVSIMC facilitates \textit{selective} shrinkage of side feature effects and promotes information sharing across the different side features. This allows BVSIMC to better isolate the side features that are most relevant for predicting drug-disease interactions, and thus achieve superior predictive performance. We verify the utility of our method through simulation studies and applications to a Mycobacterium tuburculosis drug resistance dataset and a benchmark drug repositioning dataset. In all these examples, BVSIMC is shown to outperform its competitors in terms of prediction, while also revealing the most clinically meaningful side features.

\section{Methodology}
\label{sec:method}



\textbf{Notation}: For an $r$-dimensional vector $\mathbf{x} = (x_1, \ldots, x_r)^\top$, we denote its $\ell_2$ norm as $\lVert \mathbf{x} \rVert_2$, where $\lVert \mathbf{x} \rVert_2 = \sqrt{|x_1|^2 + \cdots + |x_r|^2}$. Meanwhile, $\mathbf{0}_r$ denotes an $r$-dimensional zero vector.

\subsection{Bayesian IMC}

Let $\mathbf{Y} = (y_{ij})$ be an $I \times J$ binary matrix with labels $y_{ij} \in \{0, 1\}$, where the rows of $\mathbf{Y}$ correspond to drugs and the columns correspond to targets. An entry of $y_{ij}=1$ indicates a \textit{known} drug-disease interaction between drug $i$ and disease $j$ (i.e., resistance of the disease to the drug or a drug-disease association). In most applications, an entry of $y_{ij} = 0$ signifies \emph{either} that there is an unknown drug-target interaction \textit{or} that it is known to be absent \citep{liu2016neighborhood, zhang2020drimc}.

Let $\mathbf{u}_i \in \mathbb{R}^{d_1}$ be a vector of $d_1$ side features for the $i$th drug, and let $\mathbf{v}_j \in \mathbb{R}^{d_2}$ be a vector of $d_2$ side features for the $j$th disease. Let $\mathbf{U} = ( \mathbf{u}_1^\top, \ldots, \mathbf{u}_I^\top )^\top \in \mathbb{R}^{I \times d_1}$ and $\mathbf{V} = (\mathbf{v}_1^\top, \ldots, \mathbf{v}_J^\top)^\top \in \mathbb{R}^{J \times d_2}$ denote matrices whose rows are the side features for the drugs and the diseases respectively.

Following \citet{zhang2020drimc} and \citet{burkina2021inductive}, we map the drug and disease feature spaces onto a shared latent space through projection matrices $\mathbf{A} \in \mathbb{R}^{d_1 \times r}$ and $\mathbf{B} \in \mathbb{R}^{d_2 \times r}$ where $r$ is the user-specified column dimension. Specifically, we assume that each entry $y_{ij}$ in our data matrix is generated from a Bernoulli distribution with success probability $p_{ij} = P(y_{ij} = 1)$, where
\begin{equation} \label{logistic-factorization}
    p_{ij} = \frac{\exp(m_{ij})}{1+\exp(m_{ij})},
\end{equation}
and $m_{ij}$ denotes the $(i,j)$th entry of a latent matrix $\mathbf{M}$, where
\begin{equation} \label{M-matrix}
    \mathbf{M} = \mathbf{UA} \mathbf{B}^\top \mathbf{V}^\top.
\end{equation}
Under this logistic matrix factorization model, we follow the oversampling strategy of \cite{liu2016neighborhood} and \cite{zhang2020drimc} and introduce a confidence parameter $\xi \geq 1$ to assign greater importance to the positive cases (i.e., $y_{ij} = 1$). This practice is justified because the known drug-disease resistances or associations have been experimentlally verified, and thus, are more trustworthy and beneficial for the predictive performance of IMC \citep{liu2016neighborhood, zhang2020drimc}.

With $\xi \geq 1$, the likelihood function for $\mathbf{Y}$ is 
\begin{equation} \label{modified-likelihood}
    p \left( \mathbf{Y} \mid \mathbf{A}, \mathbf{B} \right) = \prod_{i=1}^{I}\prod_{j=1}^{J} p_{ij}^{\xi y_{ij}}\left(1-p_{ij}\right)^{1-y_{ij}} = \prod_{i=1}^{I}\prod_{j=1}^{J} \frac{\exp\left(\mathbf{u}_i^\top \mathbf{A} \mathbf{B}^\top \mathbf{v}_j\right)^{\xi y_{ij}}}{\left\{1+\exp\left(\mathbf{u}_i^\top \mathbf{A} \mathbf{B}^\top \mathbf{v}_j\right)\right\}^{\xi y_{ij} + 1 - y_{ij}}}.
\end{equation}
 Note that if the confidence parameter is set as $\xi=1$, then \eqref{modified-likelihood} reduces to the usual Bernoulli likelihood. Meanwhile, $\xi > 1$ is equivalent to duplicating the known drug-disease interactions $\xi$ times, thus assigning greater weight to these entries.

\subsection{BVSIMC Prior Formulation}

\begin{figure}[t!]
    \centering
    \includegraphics[width=0.65\textwidth]{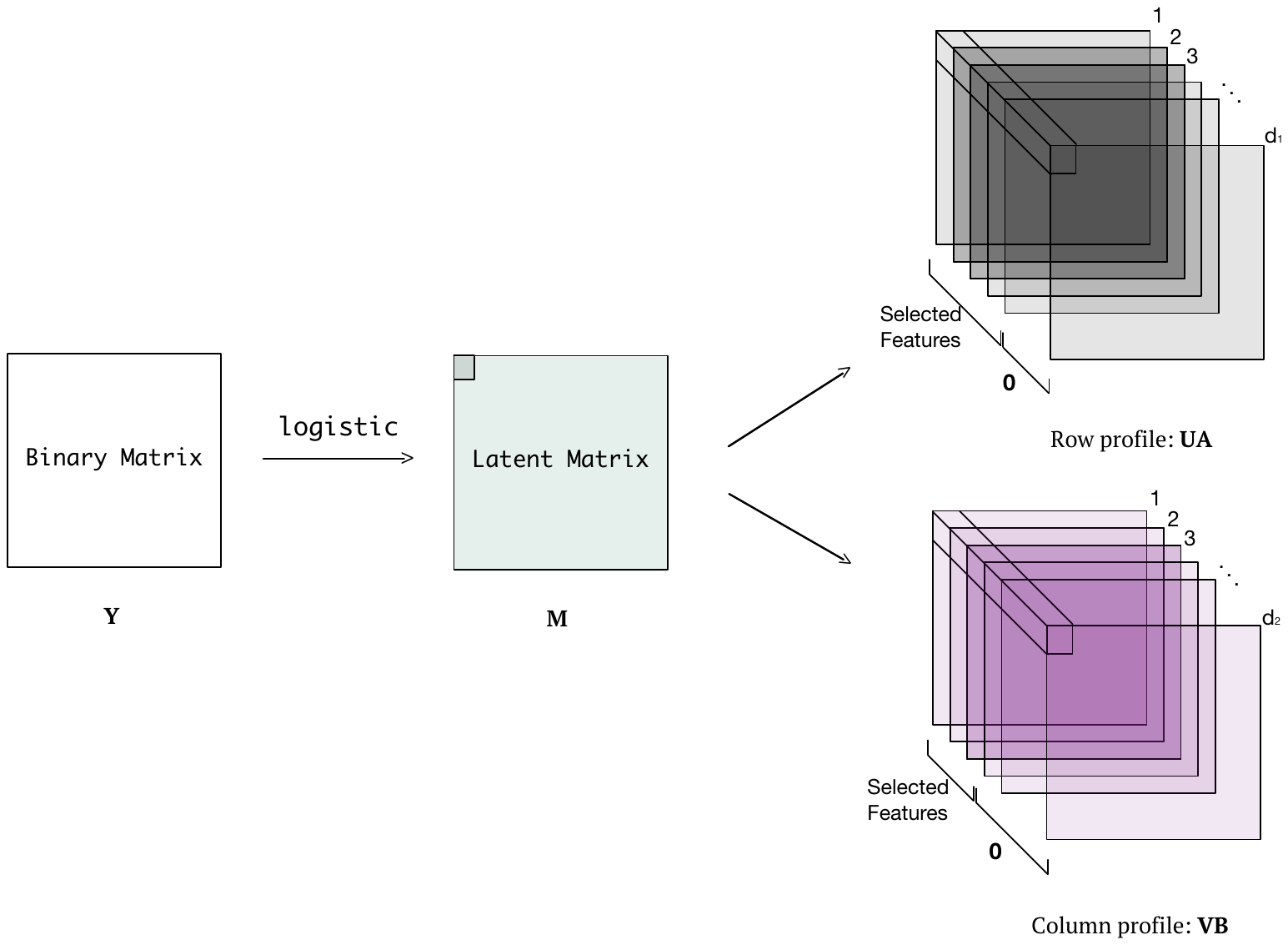}
    \caption{Overview of the proposed BVSIMC framework}
    \label{fig:BVSIMC-framework}
 \end{figure}

Under the logistic likelihood function \eqref{modified-likelihood}, we need to estimate the latent factor matrices $\mathbf{A}$ and $\mathbf{B}$. To this end, we take a Bayesian approach and endow these matrices with sparsity-inducing priors to allow for side feature selection. Figure \ref{fig:BVSIMC-framework} shows a high-level illustration of our BVSIMC framework. 

Specifically, we endow each $k$th row $\mathbf{a}_{k}, k = 1, \ldots, d_1$, in $\mathbf{A}$ and each $\ell$th row $\mathbf{b}_{\ell}, \ell = 1, \ldots, d_2$, in $\mathbf{B}$ with spike-and-slab group lasso (SSGL) priors \citep{bai2022spike}.  
\begin{equation} \label{SSGL-priors}
    \begin{aligned}
        &\pi(\mathbf{a}_k \mid \widetilde{\theta} )=\left(1-\widetilde\theta \right) \boldsymbol{\Psi} \left(\mathbf{a}_k \mid \widetilde{\lambda}_0 \right)+\widetilde\theta \boldsymbol{\Psi} \left(\mathbf{a}_k \mid \widetilde{\lambda}_1 \right), \\
        &\pi(\mathbf{b}_{\ell} \mid \theta)=\left(1-\theta \right) \boldsymbol{\Psi} \left(\mathbf{b}_{\ell} \mid \lambda_0 \right)+\theta \boldsymbol{\Psi} \left(\mathbf{b}_{\ell} \mid \lambda_1 \right), 
    \end{aligned}
\end{equation}
where, for a random $r$-dimensional vector $\mathbf{x}$,
\begin{align*}
    \boldsymbol{\Psi} (\mathbf{x} \mid \lambda ) \propto \lambda^{r}\exp\left(-\lambda \| \mathbf{x} \|_2\right)
\end{align*}
denotes an $r$-dimensional multivariate Laplace density with inverse scale parameter $\lambda$. A larger inverse scale $\lambda$ corresponds to a smaller variance for $\Psi(\cdot \mid \lambda)$. The SSGL prior \eqref{SSGL-priors} has achieved remarkable empirical success in a number of problems, including grouped linear regression \citep{bai2022spike}, generalized linear models \citep{bai2026bayesian}, and functional and longitudinal data analysis \citep{bai2023JMLR,ghosal2025variableselectionfixedrandom}. In this paper, we extend the use of SSGL priors to IMC.

In \eqref{SSGL-priors}, we set $\widetilde{\lambda}_0 \gg \widetilde{\lambda}_1$ and $\lambda_0 \gg \lambda_1$ so that the mixture components $\boldsymbol{\Psi}( \cdot \mid \widetilde{\lambda}_0)$ and $\boldsymbol{\Psi}( \cdot \mid \lambda_0)$ (or the ``spikes'') are heavily concentrated around $\mathbf{0}_r$, while the other mixture components $\boldsymbol{\Psi}(\cdot \mid \widetilde{\lambda}_1)$ and $\boldsymbol{\Psi}(\cdot \mid \lambda_1)$ (or the ``slabs'') are diffuse with a large variance. Under \eqref{SSGL-priors}, the spike densities model the sparsity in $\mathbf{A}$ and $\mathbf{B}$, whereas the slabs model the nonzero entries. The mixing proportions $\widetilde{\theta} \in (0,1)$ and $\theta \in (0,1)$ in \eqref{SSGL-priors} are the prior probabilities that $\mathbf{a}_k$ and $\mathbf{b}_{\ell}$ are drawn from the slabs instead of the spikes. Whereas the spike densities  $\boldsymbol{\Psi} (\cdot \mid \widetilde{\lambda}_0)$ and $\boldsymbol{\Psi} (\cdot \mid \lambda_0)$ shrink rows with small entries in $\mathbf{A}$ and $\mathbf{B}$ to zero, the slab densities $\boldsymbol{\Psi} (\cdot \mid \widetilde{\lambda}_1)$ and $\boldsymbol{\Psi} (\cdot \mid \lambda_1)$  allow rows with large coefficients to avoid being overshrunk and escape the pull of the spike. Thus, BVSIMC performs \textit{adaptive} shrinkage. This is in contrast to the SGIMC method of \cite{burkina2021inductive}, where the \textit{same} amount of regularization is applied to every entry in $\mathbf{A}$ and $\mathbf{B}$.







To enable BVSIMC to automatically learn the sparsity level from the data, we place independent beta hyperpriors on the mixing proportions $\widetilde{\theta}$ and $\theta$ in \eqref{SSGL-priors},
\begin{equation} \label{beta-priors-BVSIMC}
    \begin{aligned}
        &\widetilde{\theta} \sim \text{Beta}(\widetilde{\alpha}, \widetilde{\beta}), \\
        &\theta \sim \text{Beta}(\alpha, \beta), \\
    \end{aligned}
\end{equation}
where $(\widetilde{\alpha}, \widetilde{\beta}, \alpha, \beta)$ are fixed hyperparameters. The prior distributions \eqref{beta-priors-BVSIMC} on the mixing proportions $\widetilde{\theta}$ and $\theta$ render the SSGL priors on the rows of $\mathbf{A}$ and $\mathbf{B}$ \textit{non-separable}. That is, the hierarchical prior \eqref{SSGL-priors}-\eqref{beta-priors-BVSIMC} ensures that the rows within  $\mathbf{A}$ and $\mathbf{B}$ are a priori \textit{dependent}. This prior dependence allows BVSIMC to borrow information across the different side features and self-adapt to the true sparsity in the data. This affords BVSIMC a significant advantage over SGIMC \citep{burkina2021inductive}. Unlike BVSIMC, SGIMC applies a separable $\ell_2$ penalty to all entries of $\mathbf{A}$ and $\mathbf{B}$, which limits SGIMC's ability to share information between the different side features.

Under the BVSIMC prior \eqref{SSGL-priors}-\eqref{beta-priors-BVSIMC}, the posterior modes $\widehat{\mathbf{A}}$ and $\widehat{\mathbf{B}}$ for $\mathbf{A}$ and $\mathbf{B}$ are \textit{exactly} row-sparse. In particular, if $\widehat{\mathbf{a}}_k = \mathbf{0}_r$, where $\widehat{\mathbf{a}}_k$ is the $k$th row of $\widehat{\mathbf{A}}$, this implies that the $k$th drug side feature is not relevant for determining drug-disease interactions. On the other hand, if $\widehat{\mathbf{a}}_k$ is \textit{non}-zero, then the $k$th drug side feature contributes significantly to the drug-disease interactions. The rows $\widehat{\mathbf{b}}_{\ell}$'s of $\widehat{\mathbf{B}}$ corresponding to the \emph{disease} side features can similarly be interpreted. In short, by shrinking many of the rows in $\mathbf{A}$ and $\mathbf{B}$ under \eqref{modified-likelihood} to zero, BVSIMC squashes noisy side information which may be redundant -- or even detrimental -- to the prediction of drug-disease interactions. This prevents overfitting and improves the predictive accuracy of IMC.

In addition to interpretability, the row-wise sparsity incuded by the SSGL prior \eqref{SSGL-priors}-\eqref{beta-priors-BVSIMC} also overcomes the challenge of rotational ambiguity in our model. The matrices $\mathbf{A}$ and $\mathbf{B}$ in \eqref{logistic-factorization} are not identifiable because we can always post-multiply them by an orthogonal matrix $\mathbf{P}$ and obtain the exact same likelihood function in \eqref{modified-likelihood} (since $(\mathbf{AP})(\mathbf{BP})^\top = \mathbf{A}\mathbf{B}^\top$). Although rotational ambiguity is not an issue for prediction, it becomes a serious hindrance for interpretability when only individual elements $a_{km}$ or $b_{\ell n}$ are thresholded to zero. In this case, a rotation may cause  the $(k,m)$th entry $a_{km}$ or the $(\ell, n)$th entry $b_{\ell n }$ to change from zero to nonzero, or vice-versa. However, if a row $\mathbf{a}_k$ (resp. $\mathbf{b}_{\ell}$) is \textit{entirely} zero, then this row will \textit{remain} zero, even when $\mathbf{A}$ (resp. $\mathbf{B}$) is post-multiplied by an orthogonal matrix. Thus, row-wise sparsity ensures that relevant features can \emph{always} be identified, something which is not guaranteed if we were to only consider element-wise sparsity. 


\subsection{Implementation}

Under the hierarchical model \eqref{modified-likelihood}-\eqref{beta-priors-BVSIMC}, we aim to find the posterior modes $\widehat{\mathbf{A}}$ and $\widehat{\mathbf{B}}$ for the latent factor matrices $\mathbf{A}$ and $\mathbf{B}$. We can then predict the probabilities of drug-disease interactions as
\begin{align} \label{BVSIMC-predictions} 
    \widehat{p}_{ij} = \frac{\exp(\mathbf{u}_{i}^\top \widehat{\mathbf{A}} \widehat{\mathbf{B}}^\top \mathbf{v}_{j})}{1+\exp(\mathbf{u}_{i}^\top \widehat{\mathbf{A}} \widehat{\mathbf{B}} \mathbf{v}_{j})}.
\end{align}
Let $\pi(\mathbf{a}_k)$ and $\pi(\mathbf{b}_{\ell})$ denote the marginal priors for the rows $\mathbf{a}_k$ and $\mathbf{b}_{\ell}$ of $\mathbf{A}$ and $\mathbf{B}$ after integrating out the mixing proportions $\widetilde{\theta}$ and $\theta$ from the SSGL priors \eqref{SSGL-priors}. Under \eqref{modified-likelihood}, the log-posterior is
\begin{equation} \label{log-posterior}
    \begin{aligned}
        \mathcal{L} \left(\mathbf{A,B}\right) =& 
        \sum_{i=1}^I\sum_{j=1}^J \left[\xi \mathbf{Y} \odot \left(\mathbf{UAB^{\top}V^{\top}}\right)-\left(\xi \mathbf{Y}+1-\mathbf{Y}\right)\odot\log\left(1+\exp\left(\mathbf{UAB^{\top}V^{\top}}\right)\right)\right]_{i j} \\
        &+ \sum_{k=1}^{d_1} \log \pi(\mathbf{a}_k) + \sum_{\ell=1}^{d_2} \log \pi(\mathbf{b}_{\ell}). 
    \end{aligned}
\end{equation}
To find the posterior modes of $\mathbf{A}$ and $\mathbf{B}$, we utilize a coordinate ascent algorithm which iteratively updates each row of $\mathbf{A}$ or $\mathbf{B}$ holding all other rows fixed. 
Since the marginal priors for $\pi(\mathbf{a}_k)$ and $\pi(\mathbf{b}_{\ell})$ in \eqref{log-posterior} are non-differentiable, we employ an accelerated proximal gradient update \citep{beck2009fista} for each $\mathbf{a}_k, k= 1, \ldots, d_1$, and $\mathbf{b}_{\ell}, \ell = 1, \ldots, d_2$. These updates are formally derived in Section A.1 of the Appendix. Notably, the updates for $\mathbf{a}_k$ (Equations (A.10)-(A.11) in the Appendix) and $\mathbf{b}_{\ell}$ (Equations (A.12)-(A.13) in the Appendix) are all available in closed form. It is also worth mentioning that the updates for $\mathbf{a}_k$ and $\mathbf{b}_{\ell}$ are based on refined characterizations of the global posterior modes for $\mathbf{A}$ and $\mathbf{B}$ \citep{bai2022spike}. Therefore, despite having to navigate a highly nonconvex log-posterior \eqref{log-posterior}, our algorithm eliminates many suboptimal local modes from consideration \citep{bai2022spike}. This greatly increases the likelihood of our algorithm finding the global posterior modes for $\mathbf{A}$ and $\mathbf{B}$. 

Let $\mathbf{A}_{\setminus k}$ and $\mathbf{B}_{\setminus \ell}$ respectively denote the matrices $\mathbf{A}$ and $\mathbf{B}$ with their $k$th and $\ell$th rows removed. Even though the mixing proportions $\widetilde{\theta}$ and $\theta$ in \eqref{SSGL-priors} are marginalized out in \eqref{log-posterior}, we still require estimates of the conditional expectations $\widetilde{\theta}_k = \mathbb{E}[ \widetilde{\theta} \mid \mathbf{A}_{\setminus k}], k = 1, \ldots, d_1$, and $\theta_{\ell} = \mathbb{E} [ \theta \mid \mathbf{B}_{\setminus \ell} ], \ell = 1, \ldots, d_2$, in order to update $\mathbf{a}_k$ and $\mathbf{b}_{\ell}$ respectively (see Appendix A for details).  Thus, we also update these conditional expectations in each iteration. These updates, which are formally derived in Section A.2 of the Appendix, also have a closed form (lines 9 and 14 of Algorithm \ref{algo:BVSIMC-algorithm}). 

Given the highly non-convex log-posterior \eqref{log-posterior}, BVSIMC can be sensitive to the initializations $\mathbf{A}^{(0)}$ and $\mathbf{B}^{(0)}$. We considered two initialization strategies for $\mathbf{A}$ and $\mathbf{B}$: (i) random values from a standard normal distribution; (ii) a truncated singular value decomposition (SVD) with dimension $r$. The second strategy requires a prespecified dimension $r$, which serves as a hyperparameter to be tuned. In both the simulation studies and real-data applications, we adopted the grid search procedure of \cite{zhang2020drimc} and considered $r \in \{50, 100, 150, 200, 250, 300\}$ with the lower and upper bounds adjusted according to the dimensions of the data matrix $\mathbf{Y}$. We further initialized the conditional expectations as $\widetilde{\theta}_1^{(0)}, \ldots, \widetilde{\theta}_{d_1}^{(0)}$ and $\theta_1^{(0)}, \ldots, \theta_{d_2}^{(0)}$ all as $0.5$.

The complete BVSIMC algorithm is summarized in Algorithm \ref{algo:BVSIMC-algorithm}, where $\mathcal{L}$ denotes the log-posterior \eqref{log-posterior}, and $\mathbf{A}_{\setminus k}$ and $\mathbf{B}_{\setminus \ell}$ respectively denote the matrices $\mathbf{A}$ and $\mathbf{B}$ with the $k$th and $\ell$th rows removed. Note that since we used Nesterov's momentum \citep{beck2009fista} to update each row of $\mathbf{A}$ and $\mathbf{B}$, we set the parameter values in the first iteration ($t=1$) equal to their initial values. We then subsequently update these parameters in each $t$th iteration, $t \geq 2$.

\begin{algorithm}[htb!]
    \caption{BVSIMC Coordinate Ascent Algorithm}
    \label{algo:BVSIMC-algorithm}
    \begin{algorithmic}[1]
        \STATE {\bfseries Input:} data matrix $\mathbf{Y}$, side information matrices $\mathbf{U}$ and $\mathbf{V}$, fixed hyperparameters
        $\{ \widetilde{\lambda}_0, \widetilde{\lambda}_1, \widetilde{\alpha}, \widetilde{\beta}, \lambda_0, \lambda_1, \alpha, \beta, r, \xi \}$ and learning rate $\eta$
        \STATE {\bfseries Output:} posterior modes $\widehat{\mathbf{A}}$ and $\widehat{\mathbf{B}}$
        \STATE Initialize $\boldsymbol{\Omega}^{(0)} = \left\{\mathbf{A}^{(0)}, \mathbf{B}^{(0)}, \widetilde{\theta}_1^{(0)}, \ldots, \widetilde{\theta}_{d_1}^{(0)}, \theta_1^{(0)}, \ldots, \theta_{d_2}^{(0)} \right\}$
        \STATE Set $\boldsymbol{\Omega}^{(1)} = \boldsymbol{\Omega}^{(0)}$ and initialize counter at $t = 2$
        \WHILE{not converged}
            \STATE \algocm{Update rows of $\mathbf{A}$}
            \FOR{$k = 1, \cdots, d_1$}
                \STATE  
                $\mathbf{a}_k^{(t)} \leftarrow \argmax_{\mathbf{a}_k} \mathcal{L} \left( \mathbf{a}_k, \mathbf{A}_{\setminus k}^{(t-1)}, \mathbf{B}^{(t-1)} \right)$   ~ \slash \slash ~  (A.13) in Appendix A
            \STATE 
             $\widetilde{\theta}_k^{(t)} \leftarrow \frac{ \widetilde{\alpha} + \sum_{k=1}^{d_1} \mathbb{I}(\mathbf{a}_k^{(t)} \neq \mathbf{0}_r) }{\widetilde{\alpha} + \widetilde{\beta} + d_1} $ ~~ \qquad \qquad   \qquad  \slash \slash ~ Update $\widetilde{\theta}_k := \mathbb{E}[ \widetilde{\theta} \mid \mathbf{A}_{\setminus k} ]$ 
            \ENDFOR
            
            \STATE \algocm{Update rows of $\mathbf{B}$}
            \FOR{$\ell = 1, \cdots, d_2$}
                \STATE  
                $\mathbf{b}_{\ell}^{(t)} \leftarrow \argmax_{\mathbf{b}_{\ell}} \mathcal{L} \left( \mathbf{A}^{(t)}, \mathbf{b}_{\ell}, \mathbf{B}_{\setminus \ell}^{(t-1)} \right)$ ~~~ \slash \slash ~  (A.15) in Appendix A
            \STATE 
             $\theta_{\ell}^{(t)} \leftarrow \frac{ \alpha + \sum_{\ell=1}^{d_2} \mathbb{I}(\mathbf{b}_{\ell}^{(t)} \neq \mathbf{0}_r) }{\alpha + \beta + d_2} $  \qquad \qquad \qquad ~ \slash \slash ~ Update $\theta_{\ell} := \mathbb{E}[ \theta \mid \mathbf{B}_{\setminus \ell} ]$ 
            \ENDFOR

        \STATE Iterate $t \leftarrow t+1$
        \ENDWHILE
    \end{algorithmic}
\end{algorithm}

We recommend fixing the slab hyperparameters $(\widetilde{\lambda}_1, \lambda_1)$ in the SSGL priors \eqref{SSGL-priors} as $\widetilde{\lambda}_1 = \lambda_1 = 1$.  On the other hand, we recommend tuning the spike hyperparameters $(\widetilde{\lambda}_0, \lambda_0)$ and the column dimension $r$ from grids of candidate values. In the hyperpriors \eqref{beta-priors-BVSIMC}, we recommend fixing $\widetilde{\alpha} = \alpha = 1/r$ and $\widetilde{\beta} = \beta = 1$. For the confidence parameter $\xi$ and the learning rate $\eta$, we found that default values of $\xi=10$ and $\eta = 10^{-4}$ worked well in practice. However, these can also be tuned as needed.


\section{Results}  \label{Results}

In this section, we assess the performance of BVSIMC, compared to several other state-of-the-art methods: 
\begin{enumerate}
    \item traditional IMC \citep{yu2014large};
    \item SGIMC \citep{burkina2021inductive};
    \item DRIMC \citep{zhang2020drimc};
    \item NRLMF \citep{liu2016neighborhood}.
\end{enumerate}  
Traditional IMC (or simply IMC) applies ridge (squared $\ell_2$) penalties to $\mathbf{A}$ and $\mathbf{B}$ in \eqref{M-matrix} to prevent overfitting \citep{yu2014large}. However, IMC does not estimate sparse latent factor matrices and thus does not perform feature selection from the side features. SGIMC uses a combination of $\ell_1$ and $\ell_2$ penalties on the rows of $\mathbf{A}$ and $\mathbf{B}$. For DRIMC and NRLMF, we followed the procedures described in \citet{zhang2020drimc} and \citet{liu2016neighborhood} to create the similarity matrices that were then used in their methods. Neither DRIMC nor NRLMF performs feature selection. For the four competing methods, hyperparameters were either set at their default values or tuned from a grid of values as recommended by the authors. Finally, we note that there is no publicly available code for the DirtyIMC method of \citet{chiang2015matrix}, so we did not compare our approach to DirtyIMC. 

\subsection{Simulation Analysis}
\label{sec:sim}

We first investigated our proposed BVSIMC model \eqref{modified-likelihood}-\eqref{beta-priors-BVSIMC} on a small simulation study. We adapted the simulation settings of \cite{burkina2021inductive}. Specifically, we first generated two side feature information matrices as $\mathbf{U} \in \mathbb{R}^{800 \times 100}$ and $\mathbf{V} \in \mathbb{R}^{1600 \times 100}$, where all entries were generated from $\mathcal{N}(0, 0.005)$. Thus, there were 100 side features for the rows and 100 side features for the columns respectively. We then generated each $(i,j)$th entry of our $800 \times 1600$ binary matrix $\mathbf{Y}$ from an independent Bernoulli distribution with success probability $p_{ij}$ according to \eqref{logistic-factorization}-\eqref{M-matrix}, where the first 25 rows of the latent matrices $\mathbf{A} \in \mathbb{R}^{100 \times 25}$ and $\mathbf{B} = \mathbb{R}^{100 \times 25}$ were unit vectors $\mathbf{e}_1, \ldots, \mathbf{e}_{25}$, and the remaining rows were all zero vectors. Therefore, $\mathbf{A}$ and $\mathbf{B}$ each consisted of an upper $25 \times 25$ submatrix that was the identity matrix. This implies that only the first 25 row side features and only the first 25 column side features were relevant. 

After generating the complete data matrix $\mathbf{Y}$, we randomly chose $1\%$ of the entries to be observed. The other 99\% of entries were masked (i.e., they were all set to 0 to represent unknown interactions, even though some of these entries were in fact positive cases, or 1's). To mimic real applications where 0's could indicate either missing/unknown entries or known negative interactions, we mixed some of the observed 0's with the masked entries to create a new test set. Our goal was to determine how accurately we could predict the ground truth labels in this final test set. Note that since this was simulated data, we knew what the actual ground truth labels were. 

To ensure a fair comparison, we set $r=25$, i.e., the true column dimension of $\mathbf{A}$ and $\mathbf{B}$ for all methods. Based on a grid search of the spike hyperparameters, we found that BVSIMC captured the sparsity most effectively when $\widetilde{\lambda}_0 = \lambda_0 = 5$. For BVSIMC, DRIMC, and NRLMF, we considered a confidence parameter of $\xi \in \{1, 10\}$. 

\begin{table}[t!]
    \centering
        \caption{Simulation Results}
        \begin{tabular}{lc}
            \toprule
            Method & AUC  \\
            \midrule
            BVSIMC~($\xi=1$) & 0.631 \\
            BVSIMC~($\xi=10$) & \textbf{0.868} \\
             IMC            & 0.651  \\
            SGIMC              & 0.655  \\
            DRIMC~($\xi=1$)   & 0.551    \\
          DRIMC~($\xi=10$)  & 0.603  \\
            NRLMF~($\xi=1$)   & 0.620  \\
            NRLMF~($\xi=10$)  & 0.637  \\
            \bottomrule
        \end{tabular}  \label{tbl:sim}
    \end{table}

Table~\ref{tbl:sim} reports the Area Under the Receiver Operating Characteristic (AUC) for the different methods on the test data. We observed that BVSIMC with a confidence parameter of $\xi = 10$ had a significantly higher AUC score (0.868) than the other approaches, none of which were able to achieve an AUC over 0.70. In contrast to BVSIMC, DRIMC and NRLMF only showed a modest improvement in AUC when $\xi$ was incresased from 1 to 10. This may be because DRIMC and NRLMF do not regularize the side features. As a result, the similarity matrices used in DRIMC and NRLMF were contaminated by irrelevant side information, adversely affecting their prediction performance. Meanwhile, IMC and SGIMC do regularize the side features (although IMC does not shrink their effects to exactly zero), which explains their superior performance over DRIMC and NRLMF. However, neither IMC nor SGIMC performed as well as BVSIMC with $\xi=10$. As the only method to combine selective shrinkage with a confidence parameter, BVSIMC had much better predictive accuracy than the competing methods.




\subsection{Predicting Mycobacterium Tuberculosis Drug Resistance} \label{sec:drug-resistance}

 Multidrug-Resistant Tuberculosis (MDR-TB) is a severe, ongoing threat to global tuberculosis control efforts \citep{Mughal2025Cureus}. MDR-TB is caused by strains of Mycobacterium Tuberculosis (M. tb) and is a leading cause of antimicrobial resistance (AMR)-related deaths, with approximately 250,000 deaths globally each year \citep{Paul2018TheTO}. As a result, it is critically important to predict drug resistance to M. tb.

 Using data from the Bacterial and Viral Bioinformatics Resource Center (BV-BRC) \citep{BVBRC-2022} and following the approach of \cite{burkina2021inductive}, we constructed an M. tb drug dataset consisting of 6949 M. tb strains that were tested across 13 drugs: 
 Isoniazid (INH), Ethambutol (EMB), Rifampicin (RIF), Pyrazinamide (PZA), Streptomycin (STM), Ofloxacin (OFL), Capreomycin (CAP), Amikacin (AMK), Moxifloxacin (MOX), Kanamycin (KAN), Prothionamide (PTO), Ciprofloxacin (CIP), and Ethionamide (ETH). 

Side information for M. tb consisted of 9,684 genomic Single Nucleotide Polymorphism (SNP) features, while side information for the drugs consisted of 33 drug chemical functional groups. To extract SNP information, we followed instructions from the World Health Organization (WHO)\footnote{\url{https://github.com/GTB-tbsequencing/mutation-catalogue-2023} (May 2024)}. To prepare the drug side features, we used SMiles ARbitrary Target Specification (SMARTS) codes to define commonly used functional groups in chemoinformatics  \citep{SaldivarGonzalez2020}. Following \citet{burkina2021inductive}, we also augmented the side information matrices by appending an identity matrix to $\mathbf{U}$, i.e. $[\mathbf{U} \mid \mathbf{I}]$. This is a common technique in transductive matrix completion and collaborative filtering to allow a model to learn from both the baseline observed interactions (through $\mathbf{I}$) and the side features \citep{burkina2021inductive}.
 
We selected 10\% of the data as an out-of-bag (OOB) test set to validate the different methods. We then
selected a proportion $\rho \in \{ 0.001, 0.021, 0.041, \ldots, 0.181 \}$ of the entries in our $13 \times 6949$ data matrix to be observed data, leading to 10 different training sets. The remaining entries were masked. For BVSIMC, we tuned the spike hyperparameters $(\widetilde{\lambda}_0, \lambda_0)$ from the grid $\{ 1, 5, 10, 50, 100, 1000, 10{,}000 \}$ and the learning rate $\eta$ from $\{ 10^{-3}, 10^{-4}, \ldots, 10^{-8} \}$. Based on the results in Section \ref{sec:sim}, we fixed the confidence parameter to $\xi = 10$. Finally, we tuned the column dimension $r$ for all methods from the grid $\{ 5, 10, 13, 33 \}$.



\begin{figure}[t!]
  \centering
\includegraphics[width=.59\linewidth]{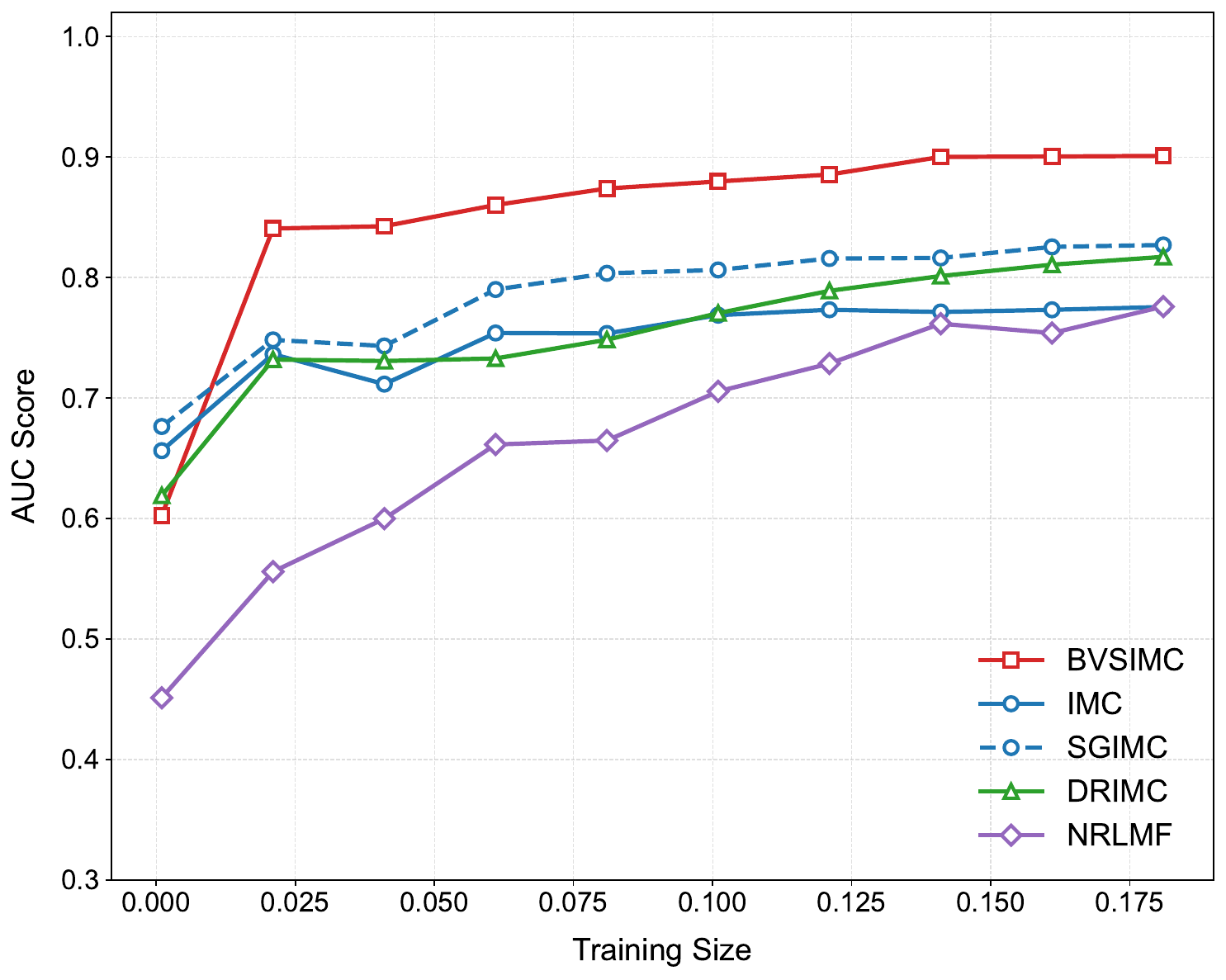}
        \caption{Results from the M. tb drug resistance prediction analysis. The training size is the proportion $\rho$ of observed (non-masked) entries.}
        \label{fig:tb-auc-comparison}
\end{figure}

Figure \ref{fig:tb-auc-comparison} shows the AUC scores for different proportions $\rho$ of observed entries. All methods improved their AUC as $\rho$ increased. However, BVSIMC consistently had the highest AUC across the greatest range of values for $\rho$. When $\rho$ was very small ($\rho=0.001$), all methods except NRLMF were comparable, with lower AUC scores of between 0.6 and 0.7. However, for all other values of $\rho$, BVSIMC had the highest AUC (above 0.8). In fact, BVSIMC was the only method to achieve an AUC greater than 0.9, once $\rho$ was 0.141 or higher.

\begin{figure}[ht]
    \centering
    \begin{subfigure}{0.43\textwidth}
        \centering
        \includegraphics[width=\linewidth]{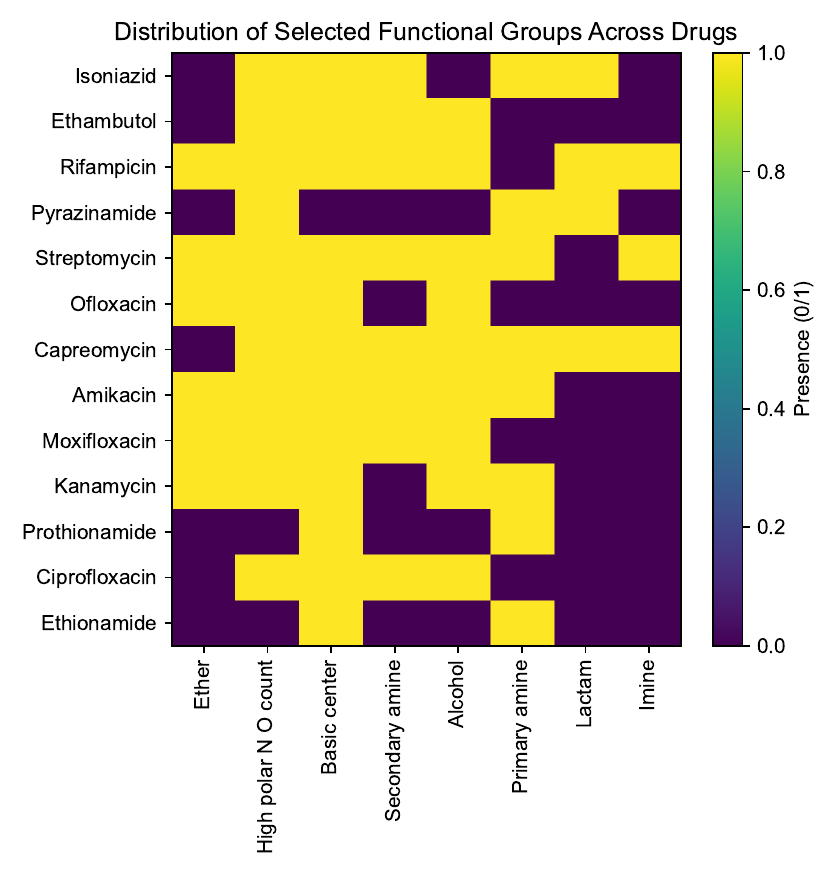}
    \end{subfigure}
    \hspace{2cm}
    \begin{subfigure}{0.42\textwidth}
        \centering
        \raisebox{0.6\height}{\includegraphics[width=\linewidth]{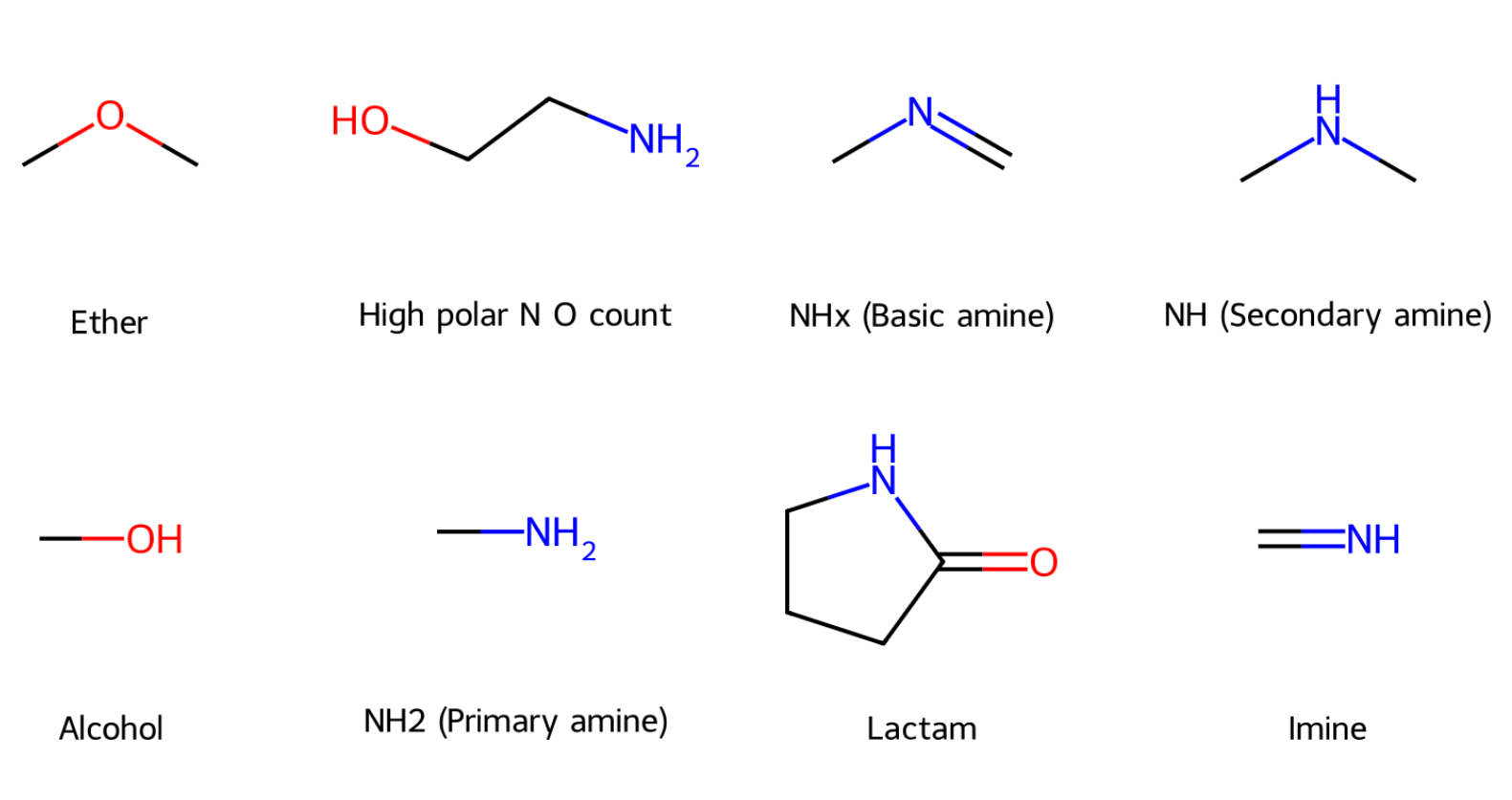}}
    \end{subfigure}
    \caption{Top eight functional groups selected by BVSIMC. Left panel: Heatmap of the 13 drugs vs. these functional groups. Right panel: Chemical structures of these groups. }
    \label{fig:tb-selection}
\end{figure}

We also examined the side features that were selected by our method for all 10 training sets. BVSIMC selected 68 SNPs and 14 drug functional groups that were highly associated with M. tb drug resistance. The left panel of Figure \ref{fig:tb-selection} displays a heatmap of the 13 drugs vs. the top eight selected functional groups. The right panel plots the chemical structures of these eight functional groups. We observed that Nitrogen and its related functional chemical groups were very popular among different drugs. BVSIMC also identified Nitrogen to be highly associated with drug resistance to M. tb. This and other selected features have the potential to aid drug design, especially compared to methods like IMC, DRIMC, and NRLMF which do not perform side feature selection.

\subsection{Predicting Drug-Target Associations in Drug Repositioning} \label{sec:drug-repositioning}

In this section, we analyze a very popular benchmarket dataset from drug repositioning research called Cdataset \citep{Luo2016DrugRB}.
While many different studies have used slightly modified versions of Cdataset, we analyzed the same version as \citet{zhang2020drimc} which contains 658 drugs and 409 diseases. In this dataset, there are 2353 known drug-disease associations. Like \citet{zhang2020drimc}, we utilized the following side features for the drugs: chemical structure, Pfam domain annotation of drug targets, and gene ontology term of targets. These were obtained from the DrugBank database \citep{wishart2018drugbank}. In total, we had 1899 side features from the drugs. For the diseases, we extracted 797 phenotype features from the Human Phenotype Ontology (HPO) database \citep{Gargano2024}, which provides a standardized vocabulary of phenotypic abnormalities encountered in human disease. The original paper by \citet{zhang2020drimc} did not use side features from HPO. While HPO provides a structured vocabulary to relate all phenotypic terms, phenotype data is still very noisy \citep{Chen2019phenotype}. Thus, we aimed to examine how performing side feature selection would affect the predictive accuracy in drug repositioning.




For BVSIMC, we tuned the spike hyperparmeters $(\widetilde{\lambda}_0, \lambda_0)$ and the learning rate $\eta$ using the same grids as those in Section \ref{sec:drug-resistance}. We also set the confidence parameter to $\xi = 10$ and tuned the column dimension $r$ from the grid $\{50, 100, 150, 200, 250, 300\}$. To compare the performance of BVSIMC to other methods, we conducted a similar experiment to that in Section \ref{sec:drug-resistance}. Namely, we designated $10\%$ of the data as an OOB test set, and we ranged the training size $\rho$ (i.e., the proportion of observed entries) from $\rho \in \{ 0.1, 0.2, \ldots, 0.8\}$. Similarly as in Section \ref{sec:drug-resistance}, we also augmented the side features matrices with identity matrices. We then fit the different IMC methods to the dataset and predicted the labels in the test set. We repeated this process 50 times.

\begin{figure}[t!]
  \centering
\includegraphics[width=.59\linewidth]{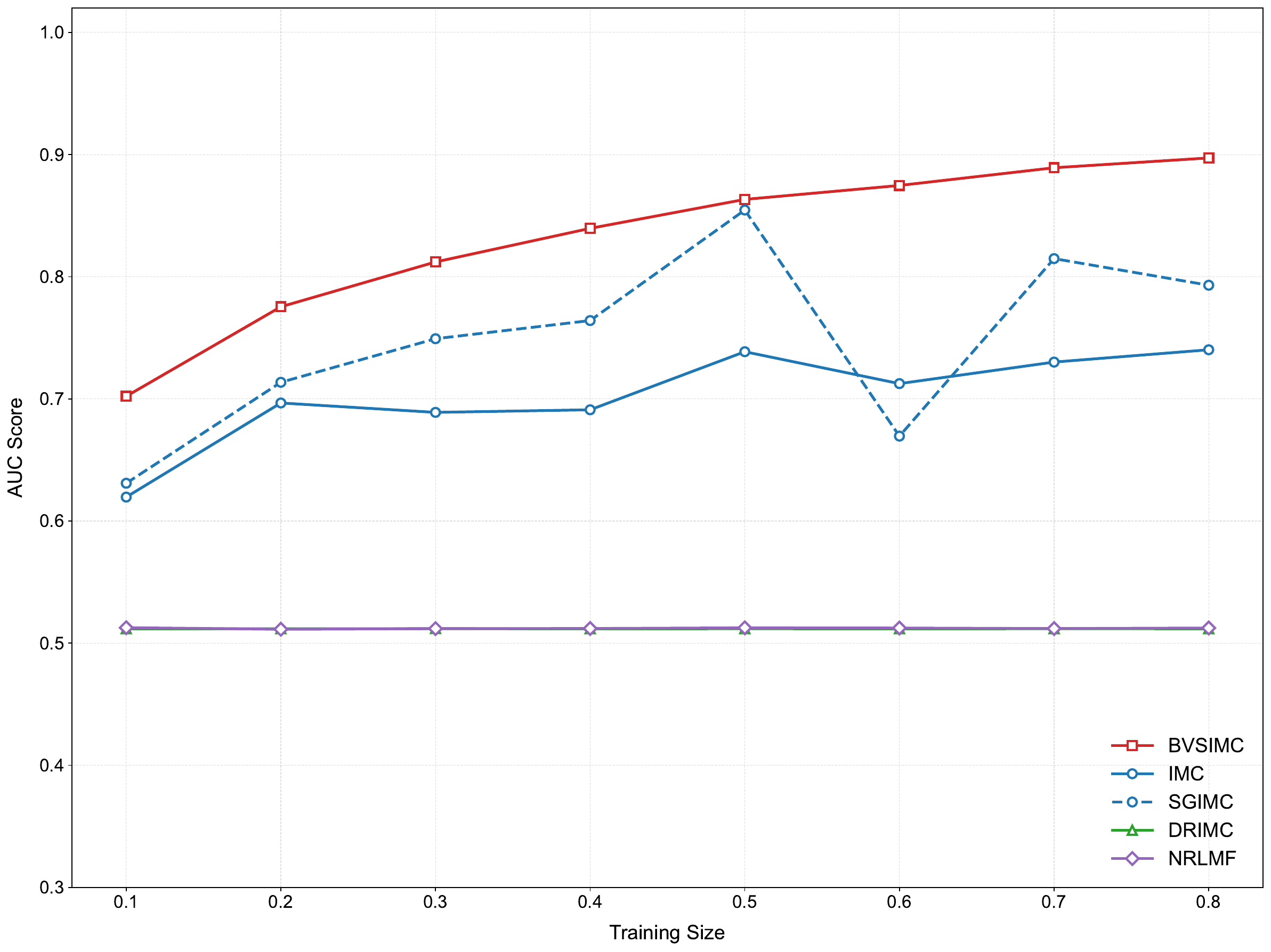}
        \caption{Drug repositioning prediction results for Cdataset. The training size is the proportion $\rho$ of observed (non-masked) entries.}
        \label{fig:cdataset-auc-noisy}
\end{figure}

Figure \ref{fig:cdataset-auc-noisy} summarizes the predictive accuracy of BVSIMC and the competing methods averaged across 50 repetitions. For all training sizes, BVSIMC had the highest AUC on average. SGIMC and IMC performed worse on average than BVSIMC but were still much better than DRIMC or NRLMF. In particular, the AUC scores were very low (close to 0.5) for DRIMC and NRLMF across all training sizes. Notably, even when we increased the confidence parameter $\xi$ for DRIMC and NRLMF, they showed little improvement in AUC. This suggests that many of the side features were noisy and actually detrimental to prediction. By regularizing the side features, BVSIMC, SGIMC, and IMC significantly outperformed DRIMC and NRLMF. However, BVSIMC still achieved the highest AUC across all training sets. We believe this is a result of both BVSIMC's selective shrinkage properties and the fact that BVSIMC assigns greater weight to known interactions. These combined characteristics make BVSIMC the most resilient to noisy side information.


\begin{figure}[t!]
    \centering
    \includegraphics[width=0.5\linewidth]{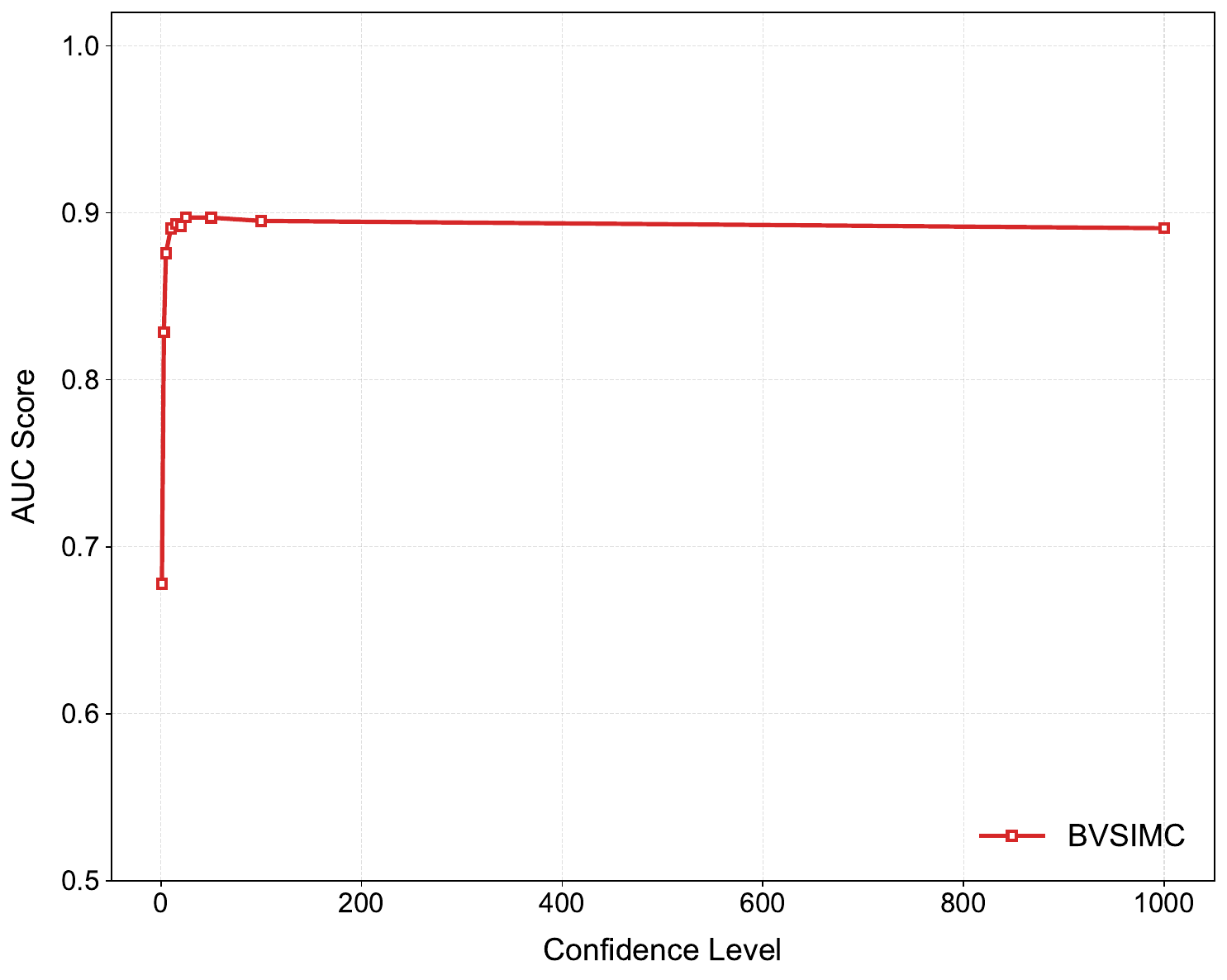}
    \caption{AUC scores under different confidence parameters $\xi$.}
    \label{fig:cdataset-auc-xi}
\end{figure}

We also examined the role of the confidence parameter $\xi$. For the training set with $\rho = 0.8$ observed entries, we ranged $\xi \in \{1, 2, 3, \ldots, 10, 50, 100, 1000 \}$ Figure \ref{fig:cdataset-auc-xi} plots the confidence level vs. the AUC score. We observe that the AUC significantly increases when $\xi$ increases from 1 to 10, but then the performance largely plateaus. This demonstrates the benefits of assigning greater weight to known interactions (effectively duplicating the positive group $\xi$ times). These  interactions have been manually verified, making them generally more trustworthy \citep{liu2016neighborhood, zhang2020drimc}. At the same time, there does not seem to be much added benefit to setting a very large $\xi$.


Overall, BVSIMC selected 838 of the 1899 drug side features and 523 of the 797 disease features. This warrants greater investigation and has the potential to help researchers better understand the main drivers of drug-disease interactions. Table \ref{tab:top10_predictions} lists the top 10 predicted drug-disease associations (in terms of predicted probability of an interaction) for drug-disease pairs $y_{ij}$ that were originally $y_{ij} = 0$ in Cdataset. These findings are also of practical interest for pharmaceutical companies and regulatory agencies.



\begin{table}[ht]
\centering
\caption{Top 10 predicted new drug–disease associations. The original entries in Cdataset were zeros for these specific drug–disease pairs.}
\label{tab:top10_predictions}
\resizebox{\textwidth}{!}{
\begin{tabular}{lllr}
\hline
\textbf{Rank} & \textbf{Drug} & \textbf{Disease} & \textbf{Probability} \\
\hline
1  & Triamcinolone & congenital adrenal hyperplasia & 0.9860 \\
2  & Busulfan & chronic myelogenous leukemia & 0.9835 \\
3  & Doxorubicin & chronic myelogenous leukemia  & 0.9800 \\
4  & Methotrexate  & chronic myelogenous leukemia & 0.9741 \\
5  & Ceftriaxone  & pulmonary fibrosis & 0.9680 \\
6  & Ceftriaxone & pulmonary fibrosis with emphysema  & 0.9680 \\
7  & Methylprednisolone  & congenital adrenal hyperplasia  & 0.9542 \\
8  & Dexamethasone & multiple myeloma & 0.9497 \\
9  & Prednisone & congenital adrenal hyperplasia & 0.9425 \\
10 & Codeine & pulmonary fibrosis with emphysema & 0.9421 \\
\hline
\end{tabular}
}
\end{table}

\section{Conclusion}
\label{sec:conclusion}

In this paper, we proposed BVSIMC, a new Bayesian variable selection-guided IMC model for binary data in drug discovery. Working within a logistic matrix fractorization framework, BVSIMC shrinks negligible rows of the latent factor matrices to zero, thus preventing overfitting and removing irrelevant side features from the fitted model. At the same time, BVSIMC performs \textit{adaptive} shrinkage, preventing overshrinkage of the most relevant side features. Through a non-separable beta-Bernoulli prior, BVSIMC allows information to be shared across different side features and self-adapts to ensemble information about sparsity. In simulation studies and two applications in drug resistance prediction and drug repositioning, we demonstrated the superior predictive performance of BVSIMC over other methods that either do not utilize or regularize side features or that do not perform adaptive shrinkage. For these drug discovery applications, BVSIMC revealed side features that were the most clinically relevant for predicting drug-target interactions.

In this paper, we focused on applications with binary data. It may be of interest to extend BVSIMC to analyses of other types of discrete data, e.g., count data in mutational signature analysis where high-dimensional genomic covariates could impact the count frequencies of specific mutation types \citep{zito2025poisson}. In addition, BVSIMC assumes that the latent matrix \eqref{M-matrix} is linear with respect to the side features. Although this worked well in the datasets we analyzed, the linearity assumption may be too restrictive, and it is of interest to capture nonlinear, low-dimensional relationships between the side features and the drug-target interactions. BVSIMC could be extended to nonlinear IMC by using kernel methods \citep{Gonen2013KBMF, Zhou2012SDM} with sparse regularization.

\section*{Competing Interests}
No competing interest is declared.

\section*{Author Contributions Statement}
Sijian Fan (Methodology [lead], Software [lead], Data curation [lead], Writing -- original draft [lead], review \& editing [supporting]), Liyan Xiong (Data curation [supporting], Writing -- review \& editing [supporting]), Dayuan Wang (Data curation [supporting], Writing -- reivew \& editing [supporting]),
Guoshuai Cai (Supervision [supporting], Writing -- review \& editing [supporting]),
Ray Bai (Supervision [lead], Methodology [supporting], Writing -- review \& editing [lead]) 

\section*{Acknowledgments}
Research reported in this publication was supported by the University of South Carolina (USC) Big Data Health Science Center, under Award Number BDHSC-2023-06. The content is solely the responsibility of the authors and does not necessarily represent the official views of the USC Big Data Science Center. The authors also thank Drs. Brian Habing, Shanghong Xie, and Jiajia Zhang for their helpful comments on a preliminary draft of this work. 

\section*{Declaration of Generative AI in Scientific Writing}
During the preparation of this work, the authors utilized ChatGPT to assist with grammar checks. After using this tool, the authors carefully reviewed and edited the content as necessary. The authors take full responsibility for the content of this article.






\bibliography{paper}

\newpage

\appendix

\numberwithin{equation}{section} 

\renewcommand{\thefigure}{A.\arabic{figure}}
\setcounter{figure}{0} 

\section{Details and Derivations for the BVSIMC algorithm}

\subsection{Updates for the Latent Factor Matrices}

Here, we derive the proximal gradient update for $\mathbf{a}_k$ (i.e., the $k$th row of $\mathbf{A}$) in detail, holding all other rows of $\mathbf{A}$ and $\mathbf{B}$ fixed at their current values. The update for the $\ell$th row $\mathbf{b}_{\ell}$ of $\mathbf{B}$ is very similar.

First, note that the marginal prior $\pi(\mathbf{a}_k)$ is defined as
\begin{equation*}
    \pi (\mathbf{a}_k) = \int_{0}^{1} \left[ (1-\widetilde{\theta}) \boldsymbol{\Psi}(\mathbf{a}_k \mid \widetilde{\lambda}_0) + \widetilde{\theta} \boldsymbol{\Psi}(\mathbf{a}_k \mid \widetilde{\lambda}_1) \right] \pi(\widetilde{\theta}) d \widetilde{\theta}, 
\end{equation*}
where $\pi(\widetilde{\theta})$ is the prior on $\widetilde{\theta}$ in (5). The log prior $\log \pi(\mathbf{a}_k)$ can be thought of as a penalty function on $\mathbf{a}_k$. Following \citet{bai2022spike}, we center this penalty so that $\text{pen}(\mathbf{0}_r) = 0$, i.e.,
\begin{align} \label{penalty-A-row}
    \text{pen}(\mathbf{a}_k) = \log \frac{\pi(\mathbf{a}_k)}{\pi(\mathbf{0}_r)} = -\widetilde{\lambda}_1 \lVert \mathbf{a}_k \rVert_2 + \log \left[ \frac{p^{\star} (\mathbf{0}_r ; \widetilde{\theta}_k, \widetilde{\lambda}_0, \widetilde{\lambda}_1)}{p^{\star} ( \mathbf{a}_k ; \widetilde{\theta}_k, \widetilde{\lambda}_0, \widetilde{\lambda}_1 ) }  \right],
\end{align}
where the function $p^{\star}(\cdot~;~\cdot, \cdot, \cdot)$ is defined as 
\begin{align} \label{pstar}
    p^{\star}(\mathbf{x}; \theta, \xi_0, \xi_1 ) = \theta \boldsymbol{\Psi}( \mathbf{x} \mid \xi_1) / [ (1-\theta) \boldsymbol{\Psi}(\mathbf{x} \mid \xi_0) + \theta \boldsymbol{\Psi}(\mathbf{x} \mid \xi_1) ],
\end{align}
and
\begin{align} \label{thetak}
    \quad \widetilde{\theta}_k = \mathbb{E} \left[\widetilde{\theta} \mid \mathbf{A}_{\setminus k}  \right].  
\end{align}
If we take the derivative of the penalty function \eqref{penalty-A-row} with respect to $\lVert \mathbf{a}_k \rVert_2$, we obtain
\begin{equation} \label{penalty-derivative}
    \frac{\partial \text{pen}(\mathbf{a}_k)}{\partial \lVert \mathbf{a}_k \rVert_2} = -\lambda^{\star} ( \mathbf{a}_k ; \widetilde{\theta}_k, \widetilde{\lambda}_0, \widetilde{\lambda}_1 ),
\end{equation}
where the function $\lambda^{\star}(\cdot ~ ;~ \cdot, \cdot, \cdot)$ is defined as
\begin{equation} \label{lambda-star}
    \lambda^{\star}(\mathbf{x}, \theta, \xi_0, \xi_1) = \xi_1 p^{\star}(\mathbf{x} ; \theta, \xi_0, \xi_1) + \xi_0 \left[ 1 - p^{\star} ( \mathbf{x} ; \theta, \xi_0, \xi_1 ) \right].
\end{equation}
Thus, we can alternatively express the SSGL penalty function \eqref{penalty-A-row} as
\begin{equation} \label{SSGL-penalty-new}
     \text{pen}( \mathbf{a}_k) = - \lambda^{\star} (\mathbf{a}_k ; \widetilde{\theta}_k, \widetilde{\lambda}_0, \widetilde{\lambda}_1) \lVert \mathbf{a}_k \rVert_2.
\end{equation}
It is worth examining the SSGL penalty \eqref{SSGL-penalty-new} in detail. Based on \eqref{lambda-star}, the amount of penalization is \emph{not} the same for each row $\mathbf{a}_k, k = 1, \ldots, d_1$, in $\mathbf{A},$. Rather, the penalty is \emph{different} for different rows of $\mathbf{A}$ depending on their magnitudes. As discussed in \citet{bai2022spike}, if $\lVert \mathbf{a}_k \rVert_2$ is small, then $\lambda^{\star} (\mathbf{a}_k ; \widetilde{\theta}_k, \widetilde{\lambda}_0, \widetilde{\lambda}_1)$ will be large (i.e.,  \textit{more} penalization will be applied to shrink $\mathbf{a}_k$ to the zero vector). On the other hand, if $\lVert \mathbf{a}_k \rVert_2$ is large, then $\lambda^{\star} (\mathbf{a}_k ; \widetilde{\theta}_k, \widetilde{\lambda}_0, \widetilde{\lambda}_1)$ will be small, i.e., \emph{less} penalization will be applied to $\mathbf{a}_k$. This affords the SSGL the property of \textit{selective} shrinkage. Instead of regularizing all rows of $\mathbf{A}$ and $\mathbf{B}$ by the same amount (as in several other competing methods), SSGL shrinks the individual rows of $\mathbf{A}$ and $\mathbf{B}$ \emph{adaptively} based on their magnitudes.

If all other rows in $\mathbf{A}$ and $\mathbf{B}$ are fixed at their current values, then based on \eqref{SSGL-penalty-new}, maximizing the log-posterior (7) with respect to $\mathbf{a}_k$ is equivalent to minimizing the objective function,
\begin{equation} \label{decomposition-obj}
    \widehat{\mathbf{a}}_k = \argmin_{\mathbf{a}_k} f(\mathbf{a}_k) - \text{pen}(\mathbf{a}_k),
 \end{equation}
 where $f(\mathbf{a}_k)$ is the differentiable negative log-likelihood function, i.e.,
\begin{equation} \label{negative-log-likelihood}
   f(\mathbf{a}_k) = \sum_{i=1}^I\sum_{j=1}^J \left[\xi \mathbf{Y} \odot \left(\mathbf{UAB^{\top}V^{\top}}\right)-\left(\xi \mathbf{Y}+1-\mathbf{Y}\right)\odot\log\left(1+\exp\left(\mathbf{UAB^{\top}V^{\top}}\right)\right)\right]_{i j} 
\end{equation}
 while $\text{pen}(\mathbf{a}_k)$ is the \textit{non}-differentiable SSGL penalty \eqref{SSGL-penalty-new}.

The decomposition of the objective \eqref{decomposition-obj} as the sum of a differentiable and a non-differentiable function of $\mathbf{a}_k$ suggests that we can use a proximal gradient descent step \citep{beck2009fista} to update $\mathbf{a}_k$. Given a learning rate $\eta>0$, the proximal gradient descent update for $\mathbf{a}_k$ at the $t$th iteration of the coordinate ascent algorithm is given by
\begin{equation} \label{proximal-gradient-update}
    \begin{aligned}
        \mathbf{a}_k^{(t)} & = \operatorname{prox}_\eta \left( \mathbf{a}_k^{(t-1)} - \eta \nabla f (\mathbf{a}_k^{(t-1)}) \right) \\
        & = \argmin_{\mathbf{x}} \ \frac{1}{2 \eta} \bigg\| \mathbf{x} - \left(\mathbf{a}_k^{(t-1)} - \eta \nabla f(\mathbf{a}_k^{(t-1)}) \right) \bigg\|_2^2 + \lambda^{\star}(\mathbf{x} ; \widetilde{\theta}_k, \widetilde{\lambda}_0, \widetilde{\lambda}_1) \lVert \mathbf{x} \rVert_2,
    \end{aligned}
\end{equation}
where $\nabla f(\mathbf{a}_k)$ is the gradient of \eqref{negative-log-likelihood} with respect to $\mathbf{a}_k$, i.e.,
\begin{equation} \label{gradient-f}
    \nabla f(\mathbf{a}_k) = \left[ \mathbf{u}_k^\top (\mathbf{W} - \xi \mathbf{Y})\mathbf{VB} \right]^\top,
\end{equation}
where $\mathbf{u}_k$ is the $k$th row of the side features matrix $\mathbf{U}$, and $\mathbf{W} = (w_{ij})$ is a matrix whose $(i,j)$th entry is $w_{ij} = (1 + \xi y_{ij} - y_{ij}) / [1 + \exp(-\mathbf{u}_i^\top \mathbf{A} \mathbf{B}^\top \mathbf{v}_j)]$.

Owing to the presence of the term $\lambda^{\star}(\mathbf{x} ; \widetilde{\theta}_k, \widetilde{\lambda}_0, \widetilde{\lambda}_1)$ in  \eqref{proximal-gradient-update}, the proximal operator is a nonconvex function of $\mathbf{x}$. In order to solve the proximal operator \eqref{proximal-gradient-update}, we have the following Proposition which gives a refined characterization of the global mode for this proximal operator. This Proposition follows from a straightforward modification of Propositions 1 and 2 and Theorem 1 of \citet{bai2022spike}.
 
\begin{proposition} \label{prop:refined}
    Let $\widehat{\mathbf{a}}_k$ denote the global mode of the proximal operator \eqref{proximal-gradient-update}, and let $\widetilde{\theta}_k$ be defined as in \eqref{thetak}. Then
     $$
    \widehat{\mathbf{a}}_{k}= 
    \begin{cases} 
        \mathbf{0}_r, & \text { when } \lVert  \mathbf{z}_{k} \rVert_2 \leq \Delta, \\ 
        \left( 1 - \frac{\lambda^{\star} ( \widehat{\mathbf{a}}_k ; \widetilde{\theta}_k, \widetilde{\lambda}_0, \widetilde{\lambda}_1 )}{\lVert \mathbf{z}_k \rVert_2} \right)_{+} \mathbf{z}_k, & \text { when } \lVert \mathbf{z}_{k} \rVert_2 >\Delta,
    \end{cases}
    $$
    where $\mathbf{z}_k = \mathbf{a}_k^{(t-1)} - \eta \nabla f(\mathbf{a}_k^{(t-1)})$, $\Delta = \inf_{\mathbf{x}} \{ \lVert \mathbf{x} \rVert_2 / 2 - \eta \cdot \text{pen}(\mathbf{x} \mid \widetilde{\theta}_k)/ \lVert \mathbf{x}_2 \}$, and $x_{+} = \max \{ 0, x \}$.

      Moreover, define the function $g(\cdot~;~ \cdot, \cdot, \cdot)$ as 
      \begin{align} \label{g-function}
      g(\mathbf{x}; \theta, \xi_0, \xi_1) = [ \lambda^{\star}(\mathbf{x} ; \theta, \xi_0, \xi_1) - \xi_1 ]^2 + (2 / \eta) \log [ p^{\star}(\mathbf{x} ; \theta, \xi_0, \xi_1) ].    
      \end{align}
      When $\left(\lambda_0-\lambda_1\right)>2 / \sqrt{\eta}$ and $g\left(\mathbf{0}_r ; \widetilde{\theta}_k, \widetilde{\lambda}_0, \widetilde{\lambda}_1 \right)>0$, the threshold $\Delta$ is bounded by
    $$
    \Delta^L<\Delta<\Delta^U,
    $$
    where
    $$
    \begin{aligned}
    & \Delta^L=\sqrt{2 \eta \log \left[1 / p^{\star} ( \mathbf{0}_{r} ; \widetilde{\theta}_k, \widetilde{\lambda}_0, \widetilde{\lambda}_1 )\right]-\eta^2 d}+\eta \widetilde{\lambda}_1, \\
    & \Delta^U=\sqrt{2 \eta \log \left[1 / p^{\star} (\mathbf{0}_{r}; \widetilde{\theta}_k, \widetilde{\lambda}_0, \widetilde{\lambda}_1 )\right]}+\eta \widetilde{\lambda}_1,
    \end{aligned}
    $$
    and
    $$
    0<d<\frac{2}{\eta}-\left(\frac{1}{\eta\left(\widetilde{\lambda}_0 -\widetilde{\lambda}_1\right)}-\sqrt{\frac{2}{\eta}}\right)^2 .
$$
\end{proposition}
When the spike hyperparameter $\widetilde{\lambda}_0$ is large, $d \rightarrow 0$ and the lower bound on the threshold approaches the upper bound in Proposition \ref{prop:refined}. This yields the approximation $\Delta \approx \Delta^U$. Thus, based on Proposition \ref{prop:refined}, we arrive at the following refined proximal update for $\mathbf{a}_k$:
\begin{equation} \label{final-update-ak}
    \mathbf{a}_k^{(t)} \leftarrow \left( 1 - \frac{\lambda^{\star}(\mathbf{a}_k^{(t-1)}, \widetilde{\theta}_k^{(t-1)}, \widetilde{\lambda}_0, \widetilde{\lambda}_1)}{\lVert \mathbf{z}_k \rVert_2} \right)_{+} \mathbf{z}_k \mathbb{I} \left( \lVert \mathbf{z}_k \rVert_2 > \Delta^{U} \right). 
\end{equation}
 It is worthwhile to point out that the refined update \eqref{final-update-ak} is a combination of soft- and hard-thresholding. This allows us to eliminate many of the suboptimal local modes for $\mathbf{a}_k$ in our algorithm through the threshold $\Delta^U$ \citep{bai2022spike}. 

To speed up convergence, we also apply a Nesterov's momentum step, akin to the fast iterative shrinkage-thresholding (FISTA) algorithm \citep{beck2009fista}. After initializing $\mathbf{a}_k^{(0)} = \mathbf{a}_k^{(1)}$, we update $\mathbf{a}_k^{(t)}$ in each $t$th iteration, $t \geq 2$, as follows:
\begin{equation} \label{ak-update-accelerated}
\boxed{
    \begin{aligned}
        & \mathbf{a}_m \leftarrow  \mathbf{a}_k^{(t-1)} + \frac{t-2}{t+1}\left(\mathbf{a}_k^{(t-1)} - \mathbf{a}_k^{(t-2)}\right), \\
        & \mathbf{z}_k \leftarrow \mathbf{a}_m - \eta \nabla f(\mathbf{a}_m), \\ 
        & \text{Update threshold } \Delta^{U} \text{ as } \\
        & \quad \quad \qquad \Delta^U = 
        \begin{cases}
    \sqrt{2 \eta \log \left[1 / p^*( \mathbf{0}_r ;  \widetilde{\theta}_{k}^{(t-1)}, \widetilde{\lambda}_0, \widetilde{\lambda}_1)\right]}+\eta \widetilde{\lambda}_1, & \text { if } g( \mathbf{0}_r ; \widetilde{\theta}_k^{(t-1)}, \widetilde{\lambda}_0, \widetilde{\lambda}_1) > 0, \\ 
    \eta \lambda^*( \mathbf{0}_r ; \widetilde{\theta}_k^{(t-1)}, \widetilde{\lambda}_0, \widetilde{\lambda}_1), & \text { otherwise}, 
\end{cases} \\
        & \text{Update } \mathbf{a}_k^{(t)} \text{ as in } \eqref{final-update-ak}. 
    \end{aligned}
    }
\end{equation}
In the complete update \eqref{ak-update-accelerated}, $\nabla f(\cdot)$ is the gradient defined in \eqref{gradient-f}, $p^{\star}(\cdot ~;~ \cdot, \cdot, \cdot)$ is defined as in \eqref{pstar}, $\lambda^{\star}(\cdot ~;~ \cdot, \cdot, \cdot)$ is defined as in \eqref{lambda-star}, and $g(\cdot ~;~ \cdot, \cdot, \cdot)$ is defined as in \eqref{g-function}.

The update for the $\ell$th row $\mathbf{b}_{\ell}$ of $\mathbf{B}$, holding all other rows in $\mathbf{A}$ and $\mathbf{B}$ fixed, is analogous to the update for $\mathbf{a}_k$ in \eqref{ak-update-accelerated}. Namely, we have the following refined proximal update for $\mathbf{b}_{\ell}$:
\begin{equation} \label{final-update-bl}
    \mathbf{b}_{\ell}^{(t)} \leftarrow \left( 1 - \frac{\lambda^{\star}(\mathbf{b}_{\ell}^{(t-1)}, \theta_{\ell}^{(t-1)}, \lambda_0, \lambda_1)}{\lVert \mathbf{z}_{\ell} \rVert_2} \right)_{+} \mathbf{z}_{\ell} \mathbb{I} \left( \lVert \mathbf{z}_{\ell} \rVert_2 > \Gamma^{U} \right), 
\end{equation}
where $\theta_{\ell}^{(t-1)}$ is the most recent update for the conditional expectation $\theta_{\ell} = \mathbb{E}[ \theta \mid \mathbf{B}_{\setminus \ell} ]$, and $\mathbf{z}_{\ell}$ and $\Gamma^{U}$ are defined below in \eqref{bl-update-accelerated}. Similarly as with the update for $\mathbf{a}_k$ in \eqref{final-update-ak}, the update for $\mathbf{b}_{\ell}$ in \eqref{final-update-bl} eliminates suboptimal local modes for $\mathbf{b}_{\ell}$ through the hard threshold $\Gamma^U$.

After initializing $\mathbf{b}_{\ell}^{(0)} = \mathbf{b}_{\ell}^{(1)}$, we update $\mathbf{b}_{\ell}^{(t)}$ in each $t$th iteration, $t \geq 2$, as follows:
\begin{equation} \label{bl-update-accelerated}
    \boxed{
    \begin{aligned}
        & \mathbf{b}_m \leftarrow  \mathbf{b}_{\ell}^{(t-1)} + \frac{t-2}{t+1}\left(\mathbf{b}_{\ell}^{(t-1)} - \mathbf{b}_{\ell}^{(t-2)}\right), \\
        & \mathbf{z}_{\ell} \leftarrow \mathbf{b}_m - \eta \nabla f(\mathbf{b}_m), \\ 
        & \text{Update threshold } \Gamma^{U} \text{ as } \\
        & \quad \quad \qquad \Gamma^U = 
        \begin{cases}
    \sqrt{2 \eta \log \left[1 / p^*( \mathbf{0}_r ;  \theta_{\ell}^{(t-1)}, \lambda_0, \lambda_1)\right]}+\eta \lambda_1, & \text { if } g( \mathbf{0}_r ; \theta_{\ell}^{(t-1)}, \lambda_0, \lambda_1) > 0, \\ 
    \eta \lambda^*( \mathbf{0}_r ; \theta_{\ell}^{(t-1)}, \lambda_0, \lambda_1), & \text { otherwise}, 
\end{cases} \\
        & \text{Update } \mathbf{b}_{\ell}^{(t)} \text{ as in } \eqref{final-update-bl}.
    \end{aligned}
    }
\end{equation}
In the complete update \eqref{bl-update-accelerated}, $\nabla f(\mathbf{b}_m) = [ \mathbf{v}_{\ell}^\top ( \xi \mathbf{Y}^\top - \mathbf{W}^\top ) \mathbf{UA} ]^{\top}$ where $\mathbf{v}_{\ell}$ is the $\ell$th row of the side features matrix $\mathbf{V}$, $\mathbf{W}$ is the same matrix as in \eqref{gradient-f}, $p^{\star}(\cdot ~;~ \cdot, \cdot, \cdot)$ is defined as in \eqref{pstar}, $\lambda^{\star}(\cdot ~;~ \cdot, \cdot, \cdot)$ is defined as in \eqref{lambda-star}, and $g(\cdot ~;~ \cdot, \cdot, \cdot)$ is defined as in \eqref{g-function}.

\subsection{Updates for the Sparsity Parameters} \label{App:sparsity-parameters}

The updates $\mathbf{a}_k^{(t)}$ and $\mathbf{b}_k^{(t)}$ require us to evaluate $\lambda^{*}(\mathbf{a}_{k}^{(t-1)}; {\widetilde{\theta}_{k}}^{(t-1)}, \widetilde{\lambda}_0, \widetilde{\lambda}_1)$ and $\lambda^{*}(\mathbf{b}_{\ell}^{(t-1)} ; \theta_{\ell}^{(t-1)}, \lambda_0, \lambda_1)$, where
\begin{equation} \label{theta-conditional-expectations}
    \widetilde{\theta}_{k} = \mathbb{E}[ \widetilde{\theta} \mid \widehat{\mathbf{A}}_{\setminus k}] \quad \text{and} \quad \theta_{\ell} = \mathbb{E}[ \theta \mid \widehat{\mathbf{B}}_{\setminus \ell}],
\end{equation}
and $\widehat{\mathbf{A}}_{\setminus k}$ and $\widehat{\mathbf{B}}_{\setminus \ell}$ respectively denote the estimated matrices $\widehat{\mathbf{A}}$ and $\widehat{\mathbf{B}}$ with their $k$th and $\ell$th rows removed. When $d_1$ and $d_2$ are large, \citet{bai2022spike} observed that these conditional expectations are very close to $\mathbb{E}[\widetilde{\theta} \mid \widehat{\mathbf{A}} ]$ and $\mathbb{E}[ \theta \mid \widehat{\mathbf{B}} ]$, respectively. Thus, for practical implementation, we can replace the conditional expectations in \eqref{theta-conditional-expectations} respectively with $\mathbb{E}[\widetilde{\theta} \mid \widehat{\mathbf{A}} ]$ and $\mathbb{E}[ \theta \mid \widehat{\mathbf{B}} ]$. These conditional expectations do not have closed-form expressions. However, by Lemma 3 of \citet{bai2022spike}, when the spike hyperparameter $\lambda_0$ is large, these expectations can be approximated as 
\begin{equation} \label{cond-expected-value-approximations}
    \mathbb{E}[\widetilde{\theta} \mid \widehat{\mathbf{A}}] \approx \frac{\widetilde{\alpha}+\widehat{q}_{\mathbf{A}}}{\widetilde{\alpha} + \widetilde{\beta} +d_1} \qquad \text{and} \qquad  \mathbb{E}[\theta \mid \widehat{\mathbf{B}} ] \approx \frac{\alpha + \widehat{q}_{\mathbf{B}}}{\alpha + \beta +d_2} 
\end{equation}
where $\widehat{q}_{\mathbf{A}}$ and $\widehat{q}_{\mathbf{B}}$ are respectively the estimated number of nonzero rows in $\widehat{\mathbf{A}}$ and $\widehat{\mathbf{B}}$. Using the approximations \eqref{cond-expected-value-approximations}, we can update the condional expectations in \eqref{theta-conditional-expectations} as
\begin{equation} \label{updates-thetak}
     \widetilde{\theta}_k^{(t)} \leftarrow \frac{ \widetilde{\alpha} + \sum_{k=1}^{d_1} \mathbb{I}(\mathbf{a}_k^{(t)} \neq \mathbf{0}_r) }{\widetilde{\alpha} + \widetilde{\beta} + d_1} \qquad \text{and} \qquad \theta_{\ell}^{(t)} \leftarrow \frac{ \alpha + \sum_{\ell=1}^{d_2} \mathbb{I}(\mathbf{b}_{\ell}^{(t)} \neq \mathbf{0}_r) }{\alpha + \beta + d_2}.
\end{equation}
For simplicity,  Algorithm \ref{algo:BVSIMC-algorithm} of the main article shows the updates \eqref{updates-thetak} performed each time an individual row of $\mathbf{A}$ or $\mathbf{B}$ is updated. However, in practice, there may be little change in the values of $ \widetilde{\theta}_k$ or $\theta_{\ell}$ between individual row updates -- or even between complete iterations of the algorithm.  Thus, it may be sensible to perform the updates \eqref{updates-thetak} only once every 10 iterations and to keep them fixed otherwise \citep{bai2022spike}.

\end{document}